\documentclass[letterpaper, 10 pt, conference]{ieeeconf}

\IEEEoverridecommandlockouts                              
\title{\LARGE \bf
iLQR for Piecewise-Smooth Hybrid Dynamical Systems
}

\author{Nathan J. Kong$^{1}$, George Council$^{1}$, and Aaron M. Johnson$^{1}$%
\thanks{This material is based upon work supported by the U.S. Army Research Office under grant \#W911NF-19-1-0080 and the National Science Foundation under grant \#ECCS-1924723. The views and conclusions contained in this document are those of the authors and should not be interpreted as representing the official policies, either expressed or implied, of the Army Research Office, National Science Foundation, or the U.S. Government. The U.S. Government is authorized to reproduce and distribute reprints for Government purposes notwithstanding any copyright notation herein. Corresponding author N.~J.~Kong {\tt\small njkong@andrew.cmu.edu}.}
\thanks{$^{1}$ Department of Mechanical Engineering, Carnegie Mellon University, Pittsburgh, Pennsylvania}
}

\usepackage{amsmath}
\usepackage{mathtools}
\usepackage{amssymb}
\usepackage{algorithm}
\usepackage{algpseudocode}
\usepackage{accents}
\usepackage{multirow}
\usepackage{cite}

\usepackage{graphicx}

\algdef{SE}[SUBALG]{Indent}{EndIndent}{}{\algorithmicend\ }%
\algtext*{Indent}
\algtext*{EndIndent}

\newcounter{theorem}

\newtheorem{definition}[theorem]{Definition}
\DeclareMathOperator*{\argmin}{arg\,min}
\allowdisplaybreaks

\begin{document}

\maketitle
\thispagestyle{empty}
\pagestyle{empty}

\begin{abstract}
Trajectory optimization is a popular strategy for planning trajectories for robotic systems.
However, many robotic tasks require changing contact conditions, which is difficult due to the hybrid nature of the dynamics. The optimal sequence and timing of these modes are typically not known ahead of time.
In this work, we extend the Iterative Linear Quadratic Regulator (iLQR) method to a class of piecewise-smooth hybrid dynamical systems with state jumps by allowing for changing hybrid modes in the forward pass, using the saltation matrix to update the gradient information in the backwards pass, and using a reference extension to account for mode mismatch. 
We demonstrate these changes on a variety of hybrid systems and compare the different strategies for computing the gradients.
\end{abstract}

\section{Introduction}

\begin{figure}[t]
    \centering
    \vspace{0.5em}
    \includegraphics[width = \columnwidth]{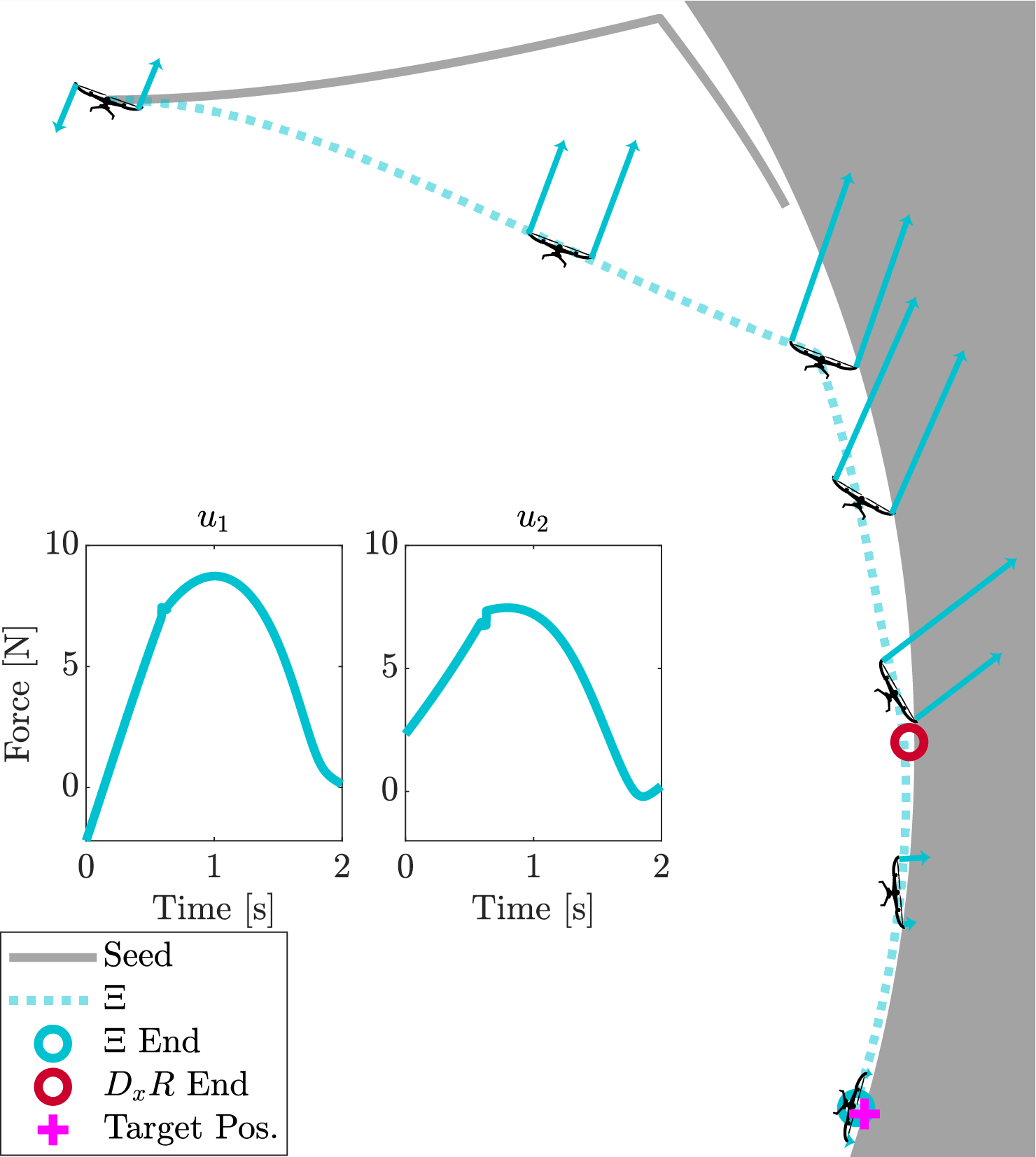}
    \caption{Demonstrating an example solution using the proposed hybrid iLQR algorithm (labeled with $\Xi$, the saltation matrix, Def.~\ref{def:salt}) where the goal is to control a quadcopter to a target final position (shown with a magenta plus) and can make contact with a curved wall with friction.
    Using a different approximation for the gradient (Jacobian of the reset map, $D_xR$, \cite{li2020hybrid}) leads to poor convergence and significantly higher cost. 
    Note that in the force plots, the optimal input is not smooth because of the hybrid transition.}
    \label{fig:quadcopter_fig}
\end{figure}

For robots to be useful in real world settings, they need to be able to interact efficiently and effectively with their environments. 
However, systems like the quadcopter perching example shown in Fig.~\ref{fig:quadcopter_fig} often have highly nonlinear dynamics and complex, time-varying environmental interactions that make trajectory planning computationally challenging. These systems are often modeled as mechanical systems with impacts, a type of hybrid dynamical system (Def.~\ref{def:hs}), \cite{Back_Guckenheimer_Myers_1993,LygerosJohansson2003,goebel2009hybrid}. Hybrid dynamical systems differ from smooth dynamical systems in many ways which make planning and control more difficult, including: 1) they contain a discrete component of state (the ``hybrid mode'') over which the continuous dynamics may differ. 2) These modes are connected by a reset function that applies a discrete (and potentially discontinuous) change to the state. 3) There may be different control authority available in each mode. 

While a wide range of trajectory optimization approaches have been proposed for smooth dynamical systems (e.g.\ \cite{betts1998survey,rao2009survey,kelly2017introduction}), most prior methods are not suitable for hybrid dynamical systems. 
One approach that has been used successfully is direct collocation, which transcribes the trajectory directly into an nonlinear program and optimizes for both the state and control input at discrete points. 
If the sequence of hybrid modes is fixed and known, the collocation can be solved as a multi-phase method \cite{von1999user,kelly2017introduction} which is a simultaneous optimization over each smooth segment with the reset map defining boundary conditions between them \cite{schultz2009modeling,posa2016optimization}.
However, the optimal mode sequence is often not known, and so contact-implicit optimization methods have been proposed \cite{posa2014direct,mordatch2012discovery}. 
These methods use complementary constraints to allow for any contact mode sequence, though such constraints are hard to solve in practice and this approach does not extend to generic hybrid systems.
Furthermore, for many real-time planning applications direct collocation methods are unfavorable because they scale poorly with time and the trajectories are not feasible until the optimization has finished.

In this paper, we propose to extend the Iterative Linear Quadratic Regulator (iLQR) method \cite{li2004iterative,tassa2012synthesis} to work for hybrid systems. iLQR (a special case of the Differential Dynamic Programming method, DDP \cite{mayne1973differential}) is a shooting method \cite{betts1998survey} that utilizes linearization in the search direction (backward pass), but implements the full nonlinear dynamics when obtaining the states of the optimized trajectory (forward pass). One advantage of iLQR, like most shooting methods, is that it can be stopped prematurely to produce a feasible trajectory \cite{posa2014direct}. 

However, traditional iLQR is defined only for smooth systems.
Here, we extend iLQR to hybrid systems by: 
    \begin{enumerate}
        \item Allowing for varying mode sequences on the forward pass by using event detection to dictate when a transition occurs and enforcing the appropriate dynamics in each mode, Sec.~\ref{section:hybrid_forwards}.
        \item Applying the reset map on the forward pass and propagating the value function through reset maps in the backwards pass by using a saltation matrix, Sec.~\ref{section:hybrid_backwards}.
        \item Using reference extensions when there is a mode mismatch to get a valid control input in each mode, Sec.~\ref{section:hybrid_extensions}.
    \end{enumerate}
In previous hybrid system DDP/iLQR work, \cite{li2020hybrid} took an important first step by extending the approach from \cite{lantoine2012hybrid} to create an ``impact aware'' iLQR algorithm which utilizes a prespecified hybrid mode sequence to allow for different dynamics and uses the Jacobian of the reset map to approximate the value function through a hybrid transition. 
Constrained dynamics and mode sequence are enforced through an Augmented Lagrangian method in an outer layer in their algorithm.
We instead use the saltation matrix (Def.~\ref{def:salt}), \cite{leine2013dynamics, rijnen2015optimal, aizerman1958determination, burden2018contraction}, to propagate the value function in the backwards pass. 
This change makes a significant difference in solution quality and convergence, as we show in Sec.~\ref{sec:results}. Furthermore, to allow use on a more general class of hybrid dynamical systems (not just rigid bodies with contact) without prespecifying the mode sequence,  the switching constraints are enforced as part of the dynamics on the forward pass -- if the current timestep reaches a hybrid event, the solution jumps to the next hybrid mode using the reset map. These changes enable the algorithm presented here to be run as a standalone algorithm with improved solution quality and convergence properties.
    

\section{Problem Formulation}
There are many different formulations of hybrid dynamical systems, with similar but subtly different properties, e.g.~\cite{Back_Guckenheimer_Myers_1993,LygerosJohansson2003,goebel2009hybrid}. For concreteness, in this paper 
we restrict our attention to the class of systems
as given in  \cite[Def.~2]{johnson2016hybrid}, with the addition of control inputs for each smooth vector field.
We elect this class as it includes mechanical systems subject to unilateral and bilateral holonomic  constraints, which are of essential interest for robotics.

\begin{definition} \label{def:hs}
    A $C^r$ \textbf{hybrid dynamical system}, for continuity class $r\in \mathbb{N}_{>0} \cup \{\infty,\omega \}$, is a tuple $\mathcal{H} := (\mathcal{J},{\mathnormal{\Gamma}},\mathcal{D},\mathcal{F},\mathcal{G},\mathcal{R})$ where the parts are defined as:
    \begin{enumerate}
        \item $\mathcal{J} := \{I,J,...,K\} \subset \mathbb{N}$ is the finite set of discrete \textbf{modes}.
        \item $\mathnormal{\Gamma} \subset \mathcal{J}\times\mathcal{J}$ is the set of discrete \textbf{transitions} forming a directed graph structure over $\mathcal{J}$.
        \item $\mathcal{D}:=\amalg_{{I}\in\mathcal{J}}$ ${D}_{I}$ is the collection of \textbf{domains} where $D_I$ is a $C^r$ manifold with corners \cite{Joyce2012}.
        \item $\mathcal{F}:= \amalg_{I\in\mathcal{J}} F_I$ is a collection of $C^r$ time-varying \textbf{vector fields} with control inputs, $F_I:  \mathbb{R}\times D_I \times U_I\to\mathcal{T}D_I$, where $U_I$ is the space of admissible control inputs in mode $I$.
        \item $\mathcal{G}:=\amalg_{(I,J)\in\mathnormal{\Gamma}}$ $G_{(I,J)}(t)$ is the collection of \textbf{guards}, where $G_{(I,J)}(t)\subset D_I \times U_I$ for each $(I,J)\in \mathnormal{\Gamma}$ is defined as a sublevel set of a $C^r$ function, i.e.\ $G_{(I,J)}(t)= \{(x,u) \in D_I \times U_I|g_{(I,J)}(t,x,u)\leq0\}$.
        \item $\mathcal{R}:\mathbb{R}\times \mathcal{G}\rightarrow \mathcal{D}$ is a $C^r$ map called the \textbf{reset} that restricts as $R_{(I,J)}:=\mathcal{R}|_{G_{(I,J)(t)}}:G_{(I,J)}(t)\rightarrow D_J$ for each $(I,J)\in \mathnormal{\Gamma}$.
    \end{enumerate}
    \end{definition}
    An execution of such a hybrid system \cite[Def. 4]{johnson2016hybrid} begins with an initial condition $x_0 \in D_I$ and input $u_I(t,x)$ and adheres to the dynamics $F_I$ on $D_I$.
    If the solution reaches guard $G_{(I,J)}$, the reset map $R_{(I,J)}$ is applied to advance the state to domain $D_J$ and activates controller $u_J(t,x)$.
    An execution is defined over a \emph{hybrid time domain} \cite[Def. 3]{johnson2016hybrid}, a disjoint union of intervals $\amalg_{{j}\in\mathcal{N}}[ \underbar{$t$}_j,\bar{t}_j]$.
    The dynamics of hybrid systems in this class can exhibit complex behavior, including sliding \cite{jeffrey2014dynamics}, branching \cite{LygerosJohansson2003}, and Zeno \cite{zhang2000dynamical}. 
    Since we make essential use of local linearization theory, we make several assumptions to eliminate these pathologies:
    \begin{itemize}
    \item Assume isolated transition surfaces with transverse \cite{hirsch1974differential, LygerosJohansson2003} intersections.
    \item Assume that there are no Zeno executions.
    \item Assume that all vector fields $F_I$ are fully controllable.
\end{itemize}

    
    Our essential tool to linearize the dynamics of a hybrid system around a switching event is the \emph{saltation matrix} \cite{leine2013dynamics, rijnen2015optimal, aizerman1958determination, burden2018contraction}, which is the necessary update for the variational equation when a hybrid transition occurs. 
    \begin{definition}[{\cite[Prop. 2]{burden2018contraction}}] \label{def:salt}
    The \textbf{saltation matrix},
    \begin{equation}
        \Xi := D_x R+\frac{\left(F_J-D_xR F_I - D_tR\right)  D_x g}{D_t g +D_x g F_I} \label{eq:saltationmatrix}
    \end{equation}
    where 
    \begin{align*}
        \Xi &:= \Xi_{(I,J)}(\bar{t}_{j},x(\bar{t}_{j}),u(\bar{t}_{j})), \,\,
        g := g_{(I,J)}(\bar{t}_j,x(\bar{t}_{j}),u(\bar{t}_{j})),\\
        R &:= R_{(I,J)}(\bar{t}_j,x(\bar{t}_{j}),u(\bar{t}_{j})), \quad
        F_I := F_I(\bar{t}_{j},x(\bar{t}_{j}),u(\bar{t}_{j})),\\
        F_J &:= F_J(\underbar{$t$}_{j+1},{R}_{(I,J)}(\bar{t}_j,x(\bar{t}_{j}),u(\underbar{$t$}_{j+1}))
    \end{align*} 
    is the first order approximation of variations at hybrid transitions from mode $I$ to $J$ and maps perturbations to first order from pre-transition $\delta x(\bar{t}_{i})$ to post-transition $\delta x(\underbar{$t$}_{i+1})$ during the $j^{th}$ transition in the following way:
    \begin{equation}
        \delta x(\underbar{$t$}_{j+1}) = \Xi_{(I,J)}\big(\bar{t}_{j},x(\bar{t}_{j})\big) \delta x(\bar{t}_{j}) + \text{h.o.t.}
        \label{eq::saltperturbation}
    \end{equation}
    where \emph{h.o.t.}\ represents higher order terms.
    \end{definition}
    
    Our class of systems are those whose executions  admit linearizations. 
        When a trajectory is away from the hybrid transition, the linearization around a trajectory $(x(t),u(t))$ is exactly the conventional variational equation $\frac{d}{dt} \delta x = D_x F_I((x(t),u(t)) \delta x + D_u F_I((x(t), u(t)) \delta u$, a linear time-varying ODE.
    At times $\bar{t}_j$ where the  trajectory $x(t)$ undergoes a discrete transition and is thus discontinuous, the variational equation is updated discontinuously with the saltation matrix.
    For a detailed description of the saltation matrix and its role in linearization, see \cite[Chp. 7]{leine2013dynamics}.
\section{Derivation/implementation}
This section introduces an abridged derivation of iLQR \cite{li2004iterative} following \cite{tassa2012synthesis}, proposes the changes to make iLQR work on hybrid systems, and discusses several important key features of the new algorithm.
\subsection{Smooth iLQR background}
Consider a nonlinear dynamical system with states $x\in\mathbb{R}^n$, inputs $u\in \mathbb{R}^m$, and dynamics $\dot{x} = F(x(t),u(t))$.
Define a discretization of the continuous dynamics over a timestep $\Delta$ such that at time $t_k$ the discrete dynamics are $x_{k+1} = f_\Delta(x_k,u_k)$, 
where $t_{k+1} =t_k+\Delta$, $x_k = x(t_k)$, and $u_k = u(t_k)$. 
Let $U := \{u_0,u_1,...,u_{N-1}\}$ be the input sequence, $J_N$ the terminal cost, and $J$ the runtime cost, where $J$ and $J_N$ are both differentiable functions into $\mathbb{R}$.

The optimal control problem over $N$ timesteps is
\begin{align} 
\min_{U} \quad & J_N(x_N) + 
      \sum_{i=0}^{N-1} J(x_i,u_i) \\ 
\text{where} \quad & x_0 = x(0)\\
& x_{i+1} = f_\Delta(x_i,u_i) \quad \forall i \in \{0, ..., N-1\} \label{eq:dynamics_constraint}
\end{align}

To solve this problem, DDP/iLQR uses Bellman recursion to find the optimal input sequence $U$, we which briefly review here.
Let $U_k:=\{u_k,u_{k+1},...,u_{N-1} \}$ be the sequence of inputs including and after timestep $k$. Define the cost-to-go $J_k$ as the cost incurred including and after timestep $k$
\begin{align}
    J_k(x_k,U_k): = J_N(x_N) + \sum_{i=k}^{N-1} J(x_i,u_i)  \label{eq:costfn}
\end{align}
with $\{x_{k+1},...,x_N\}$ the sequence of states starting at $x_k$ based on $U_k$ and \eqref{eq:dynamics_constraint}.
The value function $V$ (Bellman equation) at state $x_k$ is the optimal cost to go $J_k(x_k,U_k)$,
which can be rewritten as a recursive function of variables from the current timestep using the dynamics \eqref{eq:dynamics_constraint},
\begin{align} 
V(x_k) := & \min_{u_k} \quad J(x_k, u_k)  + V(f_\Delta(x_k,u_k)) \label{eq:substituted_value}
\end{align}
Since there is no input at the last timestep, the boundary condition of the value is  the terminal cost, $V_N(x_N) :=  J_N(x_N)$.
Next, define  $Q_k$ to be the argument optimized in \eqref{eq:substituted_value}.
Optimizing the Bellman equation directly is incredibly difficult. DDP/iLQR uses a second order local approximation of $Q_k$ where perturbations about the state and input $(x_k, u_k)$ are taken. The resulting function is defined to be
\begin{align}
    Q_k(\delta x,\delta u) := &J(x_k+\delta x, u_k+\delta u)- J(x_k, u_k) +\\
    & V(f_\Delta(x_k+\delta x,u_k+\delta u)) -V(f_\Delta(x_k,u_k)) \nonumber
\end{align}
where the value function expansion is for timestep $k+1$ and when expanded to second order
\begin{align}
    Q_k(\delta x,\delta u) \approx \frac{1}{2}\begin{bmatrix}1\\\delta x\\\delta u\end{bmatrix}^T\begin{bmatrix}
    0&Q_x^T&Q_u^T\\
    Q_x&Q_{xx}&Q_{ux}^T\\
    Q_u&Q_{ux}&Q_{uu}\\
    \end{bmatrix}\begin{bmatrix}1\\\delta x\\\delta u\end{bmatrix}\label{eq:delta_Q}
\end{align}
the expansion coefficients are
\begin{align}
    Q_{x,k}  &=  J_{x} + f_{x,k}^T V_{x}\label{eq:expansion_x}\\
    Q_{u,k}  &=  J_{u} + f_{u,k}^TV_{x}\label{eq:expansion_u}\\
    Q_{xx,k} &= J_{xx} + f_{x,k}^T V_{xx} f_{x,k} + V_{x} f_{xx,k}\label{eq:expansion_xx}\\
    Q_{ux,k} &= J_{ux} + f_{u,k}^T V_{xx} f_{x,k}+ V_{x} f_{uu,k}\label{eq:expansion_ux}\\
    Q_{uu,k} &= J_{uu} + f_{u,k}^T V_{xx} f_{u,k}+ V_{x} f_{ux,k}\label{eq:expansion_uu}
\end{align}
where subscripted variables represent derivatives of the function with respect to the variable (e.g.\ $J_x = D_x J$) and the discretized dynamics are abreviated as $f_k=f_\Delta(x_k,u_k)$.
Note that the second derivative terms (where adjacency indicates tensor contraction) with respect to the dynamics ($f_{xx,k}$, $f_{uu,k}$, and $f_{ux,k}$) in \eqref{eq:expansion_xx}--\eqref{eq:expansion_uu} are used in DDP but ignored in iLQR. 

With this value function expansion, the optimal control input, $\delta u^*$, can be found by setting the derivative of $Q(\delta x,\delta u)$ with respect to $\delta u$ to zero and solving for $\delta u$,
\begin{align}
    \delta u^* = & \argmin_{\delta u} Q(\delta x,\delta u) = -Q_{uu}^{-1}(Q_u+Q_{ux}\delta x)
\end{align}
This optimal control input can be split into a feedforward term $u_{ff} = -Q_{uu}^{-1}Q_u$ and a feedback term $K = -Q_{uu}^{-1}Q_{ux}\delta x$. Therefore, the optimal input for the local approximation at timestep $k$ is the sum of the original input and the optimal control input,
$u_k^* = u_k + \delta u^*$.

Once the optimal controller is defined, the expansion coefficients of $V$ for timestep $k$ can be updated by plugging in the optimal controller into \eqref{eq:delta_Q}
\begin{align}
    V_{x} &= Q_{x}-Q_{u}Q_{uu}^{-1}Q_{ux}\\
    V_{xx} &= Q_{xx}-Q_{ux}^TQ_{uu}^{-1}Q_{ux}
\end{align}
Now that the expansion terms for the value function at timestep $k$ can be expressed as sole a function of $k+1$ the optimal control input can be calculated recursively and stored $(u_{ff,k},K_k)$. This process is called the backwards pass.

Once the backwards pass is completed, a forward pass is run by simulating the dynamics given the new gain schedule $(u_{ff,k},K_k)$ and the previous iterations sequence of states and inputs.
\begin{align}
    \hat{x}_0 &= x_0\\
    \hat{u}_k &= K_k(\hat{x}_k - x_k) + \alpha u_{ff,k}\\
   \hat{x}_{k+1} &= f_\Delta(\hat{x}_k,\hat{u}_k) 
\end{align}
where the new trajectory is denoted with hats $(\hat{x},\hat{u})$ and $\alpha$ is used as a backtracking line-search parameters $0<\alpha\leq1$ \cite[Eqn. 12]{tassa2012synthesis}. The backwards and forwards passes are run until convergence. Following \cite{tassa2012synthesis},  convergence is when the magnitude of the total expected reduction $\delta J$ is small
\begin{align}
    \delta J(\alpha) = \sum^{N-1}_{i=0}u_{ff,i}^TQ_{u,i} + \frac{1}{2}\sum^{N-1}_{k=0}u_{ff,i}^TQ_{uu,i}u_{ff,i}
\end{align}

Convergence issues may occur when $Q_{uu}$ is not positive-definite or when the second order approximations are inaccurate. Regularization is often added to address these issues and here we use the same regularization scheme as in~\cite{tassa2012synthesis}.

\subsection{Hybrid system modifications to the forward pass}\label{section:hybrid_forwards}

The first change that is required for iLQR to work on hybrid dynamical systems is that the forward pass must accurately generate the hybrid system execution. The dynamics are integrated for the currently active mode $I_j$
for the duration of the hybrid time period $j$, i.e.\ $\forall t \in  [\underbar{$t$}_{j},\bar{t}_{j}]$, until a guard condition is met,
\begin{align}
        g_{(I_{j},I_{j+1})}(\bar{t}_{j},x(\bar{t}_{j}),u(\bar{t}_{j})) = 0
        \label{eq:guardcond}
\end{align}

To capture these hybrid dynamics in the discrete forward pass, the discretized dynamics are computed using numerical integration with event detection, so that if no event occurs the dynamic update, \eqref{eq:dynamics_constraint}, is, 
\begin{align}
    f_{\Delta_j}(\hat{x}_k,\hat{u}_k) := \int_{t_k}^{t_{k+1}} f_{I_j}(x(t),\hat{u}_k)dt + \hat{x}_k
\end{align}
If during the integration the hybrid guard condition is met, \eqref{eq:guardcond}, the integration halts, the transition state is stored, the reset map is applied, and then the integration is continued with the dynamics of the new mode, $I_{j+1}$. Suppose that the guard condition is met once (which is ensured for small times by transversality) at time $\bar{t}_{j}$, such that $t_k \leq \bar{t}_j \leq t_{k+1}$, then 
\begin{align}
    f_{\Delta}'(\hat{x}_k,\hat{u}_k) =&
    \int_{\underbar{$t$}_{j+1}}^{t_{k+1}} f_{I_{j+1}}(x(t),\hat{u}_k)dt+ \label{eq:transition_timestep}\\ 
    & \quad R_{(I_{j},I_{j+1})}\left(\bar{t}_{j},\int_{t_k}^{\bar{t}_{j}} f_{I_j}(x(t),\hat{u}_k)dt +\hat{x}_k\right)\nonumber
\end{align}
Note that this process can be repeated for as finitely many times as there are hybrid changes during a single timestep, but there cannot be infinitely many changes during a single timestep (no Zeno). In this work, we use MATLAB's ode45 method for integration and event detection. 

Finally, in addition to updating the dynamics the cost function, \eqref{eq:costfn}, can be augmented with additional cost terms, $J_{N_j}$, associated with each hybrid transition between the $M$ hybrid modes, as shown in \cite{lantoine2012hybrid},
\begin{align}
    J_0 = J_N (x_N) + \sum_{i=0}^{N-1} J(x_i, u_i) + \sum_{j=1}^{M-1} J_{N_j}(x_{N_j}) 
    \label{eq:hybridcost}
\end{align}
Such an addition may be desirable if e.g., one wanted to penalize the occurrences of a transition event in the hopes of having a minimal number of hybrid events.

\subsection{Hybrid system modifications to the backwards pass} \label{section:hybrid_backwards}
The backwards pass must be updated to reflect the discrete jumps that were added through the hybrid transitions. Away from hybrid transitions, the dynamics are smooth and behave the same way as in the smooth iLQR backwards pass, so our modification to the backwards pass occurs at timesteps where a hybrid transition is made. By substituting \eqref{eq:transition_timestep} into \eqref{eq:substituted_value}, and adding the transition cost from \eqref{eq:hybridcost}, the resulting Bellman equation for the timesteps during hybrid transition $j$ over timestep $k$ is
\begin{align} 
V(x_k) = & \min_{U_k} J(x_k, u_k)\!+\! J_{N_j}(x_{N_j})\!+\!V(f_{\Delta}'(x_k,u_k)) \label{eq:transition_bellman}
\end{align}

We elect to approximate the hybrid transition timestep to have the hybrid event occur at the end of the timestep in order to maintain smooth control inputs for each hybrid epoch.
For the backwards pass to work on the Bellman equation during transition timesteps, we need to find the linearization of $f_{\Delta}'(x_k,u_k)$. 
This linearization step is straight forward when using the saltation matrix to map perturbations pre and post hybrid transition \eqref{eq::saltperturbation}. 

The linearization can be broken up into 2 different steps, where each step the linearization is known. 
\begin{align}
    \delta x(\bar{t}_{j}) &\approx f_{x,\Delta_j}\delta x(t_{k}) + f_{u,\Delta_j} \delta u(t_{k})\\
    \delta x(\underbar{$t$}_{j+1}) &\approx \Xi\delta x(\bar{t}_{j})\label{eq:saltation_expansion}
\end{align}
where $f_{*,\Delta_{j}} = D_*f_{\Delta_j}(x,u)$ and the saltation matrix is abbreviated as $\Xi = \Xi_{(I_{j},I_{j+1})}(\bar{t}_{j},x(\bar{t}_{j}),u(t_k))$

These linearization steps can be combined and directly substituted in the coefficient expansion equations \eqref{eq:expansion_x}--\eqref{eq:expansion_uu} in place of the $f_k$ terms. For the transition cost, $J_{N_j}$, an expansion is taken about $\delta x(\bar{t}_{j})$ which can be mapped back to $(\delta x(t_k),\delta u(t_k))$ and added to the expansion coefficients. When combining all the expansion terms, the hybrid iLQR coefficients in \eqref{eq:delta_Q} are,
\begin{align}
    Q_{x,k}  &=  J_{x}+f_{x,\Delta_{j}}^T J_{x,N_j} + f_{x,\Delta_j}^T\Xi^T V_{x} \label{eq:transition_expansion_first}\\
    Q_{u,k}  &=  J_{u} +f_{u,\Delta_{j}}^T J_{x,N_j} + f_{u,\Delta_{j}}^T\Xi^T V_{x}\\
    Q_{xx,k} &= J_{xx}+f_{x,\Delta_{j}}^TJ_{xx,N_j} f_{x,\Delta_{j}}+ f_{x,\Delta_j}^T\Xi^T V_{xx} \Xi f_{x,\Delta_j}\\
    Q_{ux,k} &= J_{ux}+f_{u,\Delta_{j}}^TJ_{xx,N_j} f_{x,\Delta_{j}} + f_{u,\Delta_j}^T\Xi^T V_{xx} \Xi f_{x,\Delta_j}\\
    Q_{uu,k} &= J_{uu}+f_{u,\Delta_{j}}^TJ_{xx,N_j} f_{u,\Delta_{j}}+ f_{u,\Delta_j}^T\Xi^T V_{xx} \Xi f_{u,\Delta_j}\label{eq:transition_expansion}
\end{align}
After this update to the coefficient expansion, the backwards pass continues normally. 
If the second order variational expression for the saltation matrix is calculated, then these exact changes can be used for a hybrid DDP version of this backwards pass.
However, the computation of the second order variation expression may not be easy for systems with large state space.

As an alternative expansion, in \cite[Eq.~(21)]{li2020hybrid} the authors use an Augmented Lagrangian method to handle variations in impact time and they use the Jacobian of the reset map to propagate perturbations in state through the hybrid transition, instead of the saltation matrix \eqref{eq::saltperturbation}. 
For the hybrid backwards pass that we define, this change would be the equivalent of substituting the Jacobian of the reset map in place of the saltation matrix in \eqref{eq:saltation_expansion}
\begin{align}
\delta x(\underbar{$t$}_{j+1}) &\approx D_xR_{(I_{j},I_{j+1})}(\bar{t}_{j},x(\bar{t}_{j}),u(t_k))\delta x(\bar{t}_{j})\label{eq:reset_expansion}
\end{align}
and similarly in the hybrid coefficient expansion equations \eqref{eq:transition_expansion_first}--\eqref{eq:transition_expansion}.
We show empirically that using this alternate version with the Jacobian of the reset map does not perform as well as using the saltation matrix and may not converge.

\subsection{Hybrid extensions for mode mismatches}\label{section:hybrid_extensions}
Since the forward pass can alter the contact sequence, the new trajectory is not confined to the previous trajectory's mode sequence or timing. This feature is intended because the algorithm can now remove, add, or shift mode transitions if cost is reduced. However, this introduces an issue when the reference mode is not the same as the current mode. 

In \cite[Eq.~7]{rijnen2015optimal}, the authors consider the problem of mode mismatch for an optimal hybrid trajectory, both of the reference and of the feedback gains -- the reference is extended by integration, and the gains are held constant.
We employ their strategy, as well  as apply this same rule for the input and the feedforward gains -- applying the input intended for a different mode can cause destructive results, or be not well-defined.
If the number of hybrid transitions exceeds that of the reference, we elected to hold the terminal state and gains constant, though other choices could be made instead.


\subsection{Algorithm} \label{section:algorithm}

With each hybrid modification to iLQR listed in Sections \ref{section:hybrid_forwards}, \ref{section:hybrid_backwards}, and \ref{section:hybrid_extensions} our new algorithm can be summarized as follows: 1) Given some initial state, input sequence, quadratic loss function, number of timesteps, and timestep duration a rollout is simulated to get the initial reference trajectory and mode sequence. 2) A hybrid backwards pass (using the regularization from \cite{tassa2012synthesis}) computes the optimal control inputs for the reference trajectory. 3) Hybrid reference extensions are computed on the start and end states for each hybrid reference segment. 4) The forward pass simulates the current mode's dynamics until a hybrid guard condition is met or it is the end of the simulation time; if the guard is reached, the corresponding reset map is applied and the simulation is continued. This forward pass is repeated with a different learning rate until the line search conditions are met \cite{tassa2012synthesis}. 5) Then the backwards pass, hybrid extensions, and forward passes are repeated until convergence. 

\section{Hybrid System Examples and Experiments}
In this section, we define a set of hybrid systems -- ranging from a simple 1D bouncing ball to a perching quadcopter with constrained dynamics and friction -- and a series of experiments which evaluates how our hybrid iLQR algorithm performs in a variety of different settings. 

For all of the examples, we assume that there is no desired reference trajectory to track and that there is no hybrid transition cost $J_{N_j}$ -- this means the runtime cost is only a function of input. 
In each experiment, a comparison against using the Jacobian of the reset map instead of the saltation matrix is made by evaluating the expected cost reduction for the entire trajectory and the final cost. 
The Jacobian of the reset variant is labeled as $D_xR$-iLQR and the main variant which uses the saltation matrix $\Xi$-iLQR. 

For all examples, $m=1$ is the mass of a rigid body, $g=9.8$ is the acceleration due to gravity, the number of timesteps simulated is $N=1000$, and the timestep duration is $\Delta = 0.001$s unless specified.

The dynamics considered here fall into the category of Euler Lagrange dynamics subjected to unilateral holonomic constraints. We use the dynamics, impact law, and complementarity conditions as derived in \cite{johnson2016hybrid}.
These systems have configuration variables $q$ where the state of the system is the configurations and their time derivatives $x = [q^T,\dot{q}^T]^T$.
When the system is in contact with a constrained surface $a(q) = 0$, a constraint force $\lambda$ is applied to not allow penetration in the direction of the constraint. The accelerations $\ddot{q}$ and constraint forces $\lambda$ are found by solving the constraint and accelerations simultaneously,
\begin{align}
    M(q)\ddot{q} &+C(q,\dot{q})\dot{q} +N(q,\dot{q})+ A(q)^T\lambda= 
    \Upsilon(q,u)
    \label{eq:constrained_dyn}\\
    A(q)\ddot{q} &+ \dot{A}(q)\dot{q} = 0
\end{align}
where $M(q)$ is the manipulator inertia matrix, $C(q,\dot{q})$ are the Coriolis and centrifugal forces, $N(q,\dot{q})$ are nonlinear forces including gravity and damping, $A(q)  = D_q a(q)$ is the velocity constraint, and $\Upsilon(u)$ is the input mapping function.

Suppose the constrained surface $a_J(q)$ is the $J$th possible hybrid mode, and the current mode is the unconstrained mode. $a_J(q)$ acts as the guard surface for impacts  $g_{(1,J)} = a_J(q)$. When the system hits the impact guard, the velocity is reset using a plastic or elastic impact law \cite{johnson2016hybrid}.

Releasing a constrained mode (liftoff) occurs when a constraint force becomes attractive rather than repulsive; thus we define hybrid guard $g(t,x,u) := \lambda$ and the reset map at these events are identity transforms because no additional constraints are being added.

\subsection{Bouncing ball elastic impact}
We begin with a 1D bouncing ball under elastic impact \cite{goebel2009hybrid}, where the state $x=[z,\dot{z}]^T$ is the vertical position $z$ and velocity $\dot{z}$. 
The input $u$ is a force applied directly to the ball.
The two hybrid modes, $1$ and $2$, are defined when the ball has negative velocity $\dot{z}<0$ and when the ball has non-negative velocity $\dot{z}\geq 0$, respectively.
The dynamics on each mode are ballistic dynamics plus the input
\begin{align}
    F_1(x,u) = F_2(x,u) := \left[\dot{z},\frac{u-mg}{m}\right]^T
\end{align}
Hybrid mode $1$ transitions to $2$ when the ball hits the ground, $g_{(1,2)}(x) := z $, and mode $2$ transitions to $1$ at apex $g_{(2,1)}(x) := \dot{z}$.
When mode $1$ transitions to $2$, an elastic impact is applied, $R_{(1,2)}(x) = [z,-e \dot{z}]^T$ where $e$ is the coefficient of restitution.
The reset map from $2$ to $1$ is identity.

The Jacobian of the reset map and saltation matrix are,
\begin{align}
    D_xR_{(1,2)} = \begin{bmatrix}
    1 & 0\\
    0 & -e
    \end{bmatrix}, \, \Xi_{(1,2)} = \begin{bmatrix}
    -e & 0\\
    \frac{(u-mg)(e+1)}{m\dot{z}} & -e
    \end{bmatrix}
\end{align}
When transitioning from $2$ to $1$, both Jacobian of the reset map and saltation matrix are identity.

The problem data is to have the ball fall from an initial height with no velocity, $x_0 = [4,0]^T$, and end up at a final height $x_{des}$ with no velocity with penalties $R = 5\times10^{-7}/\Delta$, $Q_N = 100I_{2\times 2}$ and the problems were seeded with a constant input force to obtain different number of bounces.
A suite of bouncing conditions are considered and are summarized in Table \ref{table::experiment_table}. 
In the case where $0$ bounces are optimal $x_{des} = [3,0]^T$ while
where $1$ or $3$ bounces are optimal $x_{des} = [1,0]^T$. For $3$ bounces the timestep is set to $\Delta = 0.004$.
To evaluate the effectiveness of the hybrid extensions, Sec.~\ref{section:hybrid_extensions}, an additional comparison using our hybrid iLQR algorithm where we do not apply any hybrid extensions is made. 
For all cases, a convergence cutoff for this problem is set to be if $|\delta J|\leq 0.05$.

\begin{table}[tb]
\vspace{1em}
\caption{Bouncing ball with elastic impacts. Trials vary in optimal number of bounces, number of seeded bounces, which method was used, total cost, and convergence $|\delta J| < 0.05$}

\centering
\begin{tabular}{c c c c c c}
\hline
Optimal $\#$ & Seed $\#$ & Method & Actual $\#$& Cost  & Converged\\
\hline
0&0& $\Xi$ & 0 & $53.1$ & Yes\\
0&0& $D_xR$ & 0 & $53.1$ & Yes\\
0&1& $\Xi$ & 1 & $114$ & Yes\\
0&1& $D_xR$ & 0 & $53.1$ & Yes\\
0&1& Direct & 1 & $114$ & Yes\\
\hline
1&0& $\Xi$ & 0 & $97.3$ & Yes\\
1&0& $D_xR$ & 0 & $97.3$ & Yes\\
1&1& $\Xi$ & 1 & $42.5$ & Yes\\
1&1& $D_xR$ & 0 & $97.3$ & Yes\\
1&3& $\Xi$ & 1 & $42.5$ & Yes\\
1&3& $D_xR$ & 1 & $125$ & No\\
\hline
3&1& $\Xi$ & 1 & $105$ & Yes\\
3&1& $D_xR$ & 0 & $114$ & Yes\\
3&3& $\Xi$ & 3 & $0.536$ & Yes\\
3&3& $D_xR$ & 3 & $19.6$ & No\\
3&3& No Ext. & 3 & $53.3$ & No\\
\hline
\label{table::experiment_table}
\end{tabular}
\end{table}

\subsection{Ball dropping on a spring-damper}
Hard contacts are sometimes relaxed using springs and dampers, so we consider the 1D bouncing ball case, but instead of having a discontinuous event at impact, the impact event is extended by assuming the ground is a spring damper (i.e., a force law  $f_{sd}(z,\dot{z}) := kz+d\dot{z})$ when being penetrated and a spring when releasing.
The system now has an identity reset, but since the saltation matrix is not identity, the hybrid transition still produces a jump in the linearization.

The hybrid modes are defined as: the aerial phase $1$, the spring-damper phase $2$ and the spring phase $3$.
The spring and dampening coefficients are chosen to be $k = 100$ and $d = 5$.
The guards are when the ball hits the ground $g_{(1,2)} = z$, when the ball no longer has any penetrating velocity $g_{(2,3)} = \dot{z}$, and when the ball is released from the ground $g_{(3,1)} = z$.
For all of these transitions, the reset map is an identity transformation and the states do not change.

The example is setup to have the ball fall an initial height with an initial downwards velocity $x_0 = [3,-2]$, end up at a height with no velocity  $x_{des} = [1,0]$, with penalties $R = 0.0001$, $Q_N = 100 I_{2\times2}$ and no input for the seed.




\subsection{Ball drop on a curved surface with plastic impacts}
To test our algorithm with a nonlinear constraint surface, we designed a system where an actuated ball in 2D space is dropped inside a hollow tube and is tasked to end in a goal location on the tube surface. 


The configuration states of the system are the horizontal  and vertical positions $q = [y, z]^T$.
This system consists of two different hybrid modes: the unconstrained mode $1$ and in the constrained mode $2$. 
The constrained surface is defined to be a circle with radius $2$, $a(q) = 4-y^2-z^2$.
The dynamics of the system, \eqref{eq:constrained_dyn}, are ballistic dynamics with direct inputs on configurations, $M(q) = mI_{2\times2}$, $N(q,\dot{q}) =[0,-mg]^T$, $C(q,\dot{q}) = 0_{2\times2}$, and $\Upsilon=[u_y,u_z]^T$.
The impact guard from (1,2) is defined by the circle's constrained surface and the liftoff guard from (2,1) is the constraint force $\lambda$.

The example is setup to have the ball fall from an initial height with velocity pointing down and to the right $x_0 = [1,0]$, end up at a specific location on circle with no velocity  $x_{des} = [-\sqrt{3},-1,0,0]$, with penalties $R = 0.0001$, $Q_N = 100I_{4\times 4}$ and no input for the initial seed except for a vertical force $2mg$ applied for a small duration to cause the ball to momentarily leave the constraint.

\subsection{Perching quadcopter}
We introduce a quadcopter perching example inspired by \cite{lussier2011landing}, where we consider a planar quadcopter which can make contact with sliding friction on a surface. 
When both edges of the quadcopter are touching the constraint, we assume some latching mechanism engages and fully constrains the quadcopter in place with no way to release.
This problem explores planning with an underactuated system, friction, constraint surfaces, nonlinear dynamics, nonlinear guards, and nonlinear resets.

The configurations of the system are the vertical, horizontal, and angular position $q = [y,z,\theta]^T$ and the inputs are the left and right thrusters, $u_1$ and $u_2$.
The dynamics are defined by \eqref{eq:constrained_dyn} with the following 
\begin{align}
    M(q) &:= \begin{bmatrix}
    m&0&0\\
    0&m&0\\
    0&0&I
    \end{bmatrix}, \quad
    C(q,\dot{q}) := \begin{bmatrix}
    0&0&0\\
    0&0&0\\
    0&0&0
    \end{bmatrix},\\
     N(q,\dot{q}) &:= \begin{bmatrix}
    0\\
    -mg\\
    0
    \end{bmatrix}, \quad
    \Upsilon := \begin{bmatrix}
    -\sin(\theta)(u_1 + u_2)\\
    \cos(\theta)(u_1 + u_2)\\
    \frac{1}{2}(u_2w - u_1w)
    \end{bmatrix}
\end{align}
where $w = 0.25$ is the width and $I = 1$ is the inertia of the quadcopter.


To add more complex geometry, the constrained surface is a circle centered about the origin with radius 5.
Since the edges of the quadcopter make contact with the surface, the left and right edges of the quadcopter are located at,
\begin{align}
    [y_L,z_L]^T &= [y-\frac{1}{2}w\cos{\theta},z-\frac{1}{2}w\sin{\theta}]^T\\
    [y_R,z_R]^T &= [y+\frac{1}{2}w\cos{\theta},z+\frac{1}{2}w\sin{\theta}]^T
\end{align}
The constraints are then $a_1 = 25- y_L^2 - z_L^2$ and $a_2 = 25- y_R^2 - z_R^2$.
Frictional force $\lambda_t$ is defined to be tangential to the constraint with magnitude proportional to the constraint force $\lambda_n$, $\lambda_{t} = \mu \lambda_n$, where $\mu$ is the coefficient of friction. 


The example is setup to have the quadcopter start some distance away from the constraint with a horizontal velocity, $x_0 = [2,2.5,-\pi/8,4,0,0]^T$, end up oriented with the constraint with no velocity $x_{des} = [5\cos(-\pi/12),5\cos(-\pi/12),-7/12\pi,0,0,0]^T$, timesteps $\Delta = 0.002$, with penalties $R = 0.01_{2\times2}$, and $Q_N = [1000I_{3\times3},0_{3\times3}; 0_{3\times3}, 0.1I_{3\times3}]$.
The position portion is weighted more heavily than velocity because the goal is to get close enough to the desired location to perch.
For the seed, a combined thrust of equal to $1.5mg$ was applied constantly and if both edges made contact with the constraint, the thrust force was dropped to $0.1mg$. This initial input resulted in a trajectory which makes contact with the right edge and then shortly after makes double contact with the constraint as shown in Fig.~\ref{fig:quadcopter_fig}.

\section{Results}
\label{sec:results}
In this section, the results of the experiments on each system are presented. Overall, the Jacobian of the reset map method $D_xR$-iLQR has trouble converging and has worse cost compared to our proposed algorithm $\Xi$-iLQR which uses the saltation matrix.
\subsection{Bouncing Ball with Elastic Impacts}
\begin{figure}[t]
    \centering
    \vspace{0.5em}
    \includegraphics[width = \columnwidth]{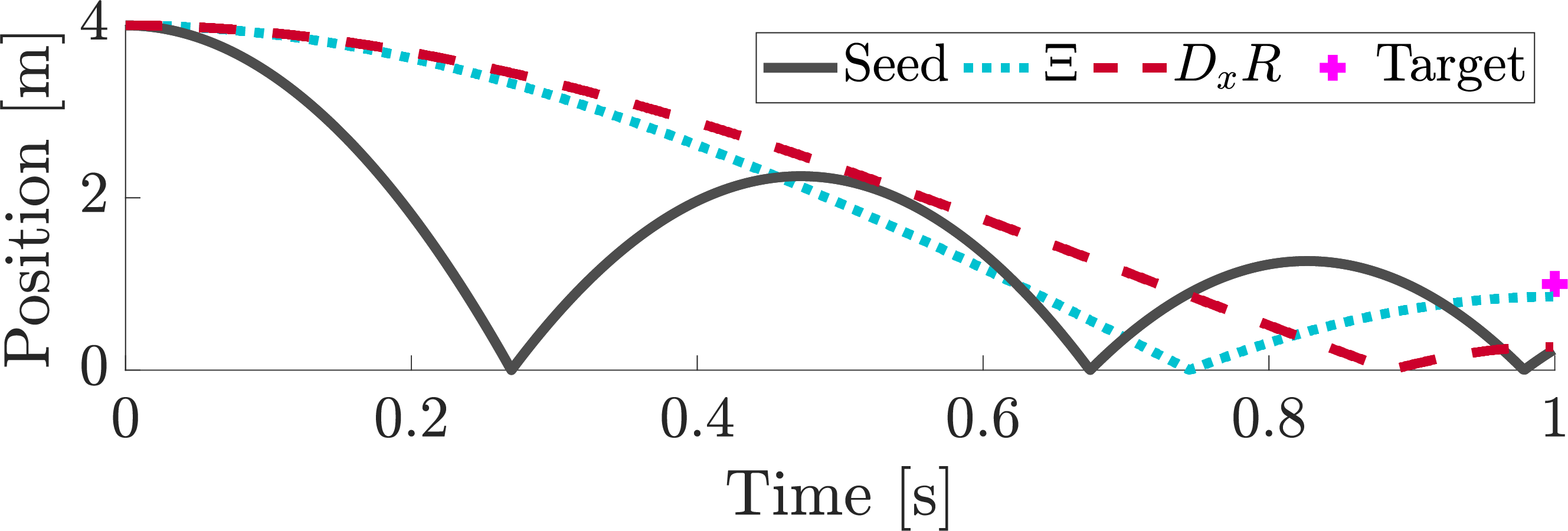}
    \caption{Bouncing ball with elastic impact where 1 bounce is optimal and 3 bounces are seeded. The target end position is shown in (magenta plus).
    Both gradient update methods were able to pull away the unnecessary bounces, but the method using $D_xR$ did not converge or get to the target state.}
    \label{fig:triple_bounce}
\end{figure}
The outcomes of the experiment comparing $D_xR$-iLQR to $\Xi$-iLQR are shown in Table \ref{table::experiment_table}. An example run is shown in Fig. \ref{fig:triple_bounce}.
$D_xR$-iLQR did not converge ($|\delta J|>0.05$) on any example if a hybrid transition was maintained, while $\Xi$-iLQR converged on every example. 
The only cases where $D_xR$-iLQR converged were when the algorithm removed all of the bounces -- 
which becomes equivalent to smooth iLQR.
$\Xi$-iLQR has lower cost compared to $D_xR$-iLQR for every example except for when the problem is seeded with no bounces (they obtain the same smooth solution) and when no bounces was the optimal solution but the problem was seeded with a single bounce. In this case, $\Xi$-iLQR did converge to a different local minima\footnote{
This solution was confirmed as a local minima under a single bounce by comparing it against a trajectory produced using direct collocation \cite{kelly2017introduction} constrained to a single bounce, as shown in Table. \ref{table::experiment_table}.
}, which is not surprising as it is not a global optimization. 

The value of the hybrid extension was tested on the three bounce optimal three bounce seeded case. Without the hybrid extension, the optimizer did not converge and did significantly worse than $D_xR$-iLQR. 
This highlights the importance of the hybrid trajectory extensions: even though the backwards pass is correct, having mode mismatches will lead to unfavorable convergence and trajectory quality.

Overall, $\Xi$-iLQR produced locally optimal solutions for each variation and was able to remove unnecessary bounces in some cases, though it never added any. 
This result is expected because there is no gradient information on the backwards pass being provided to give knowledge about adding additional bounces. Furthermore, as discussed above, there may not be an appropriate controller available when a novel hybrid mode is encountered.

\subsection{Ball dropping on a spring-damper}
    \begin{figure}[t]
    \centering
    \vspace{0.5em}
    \includegraphics[width = \columnwidth]{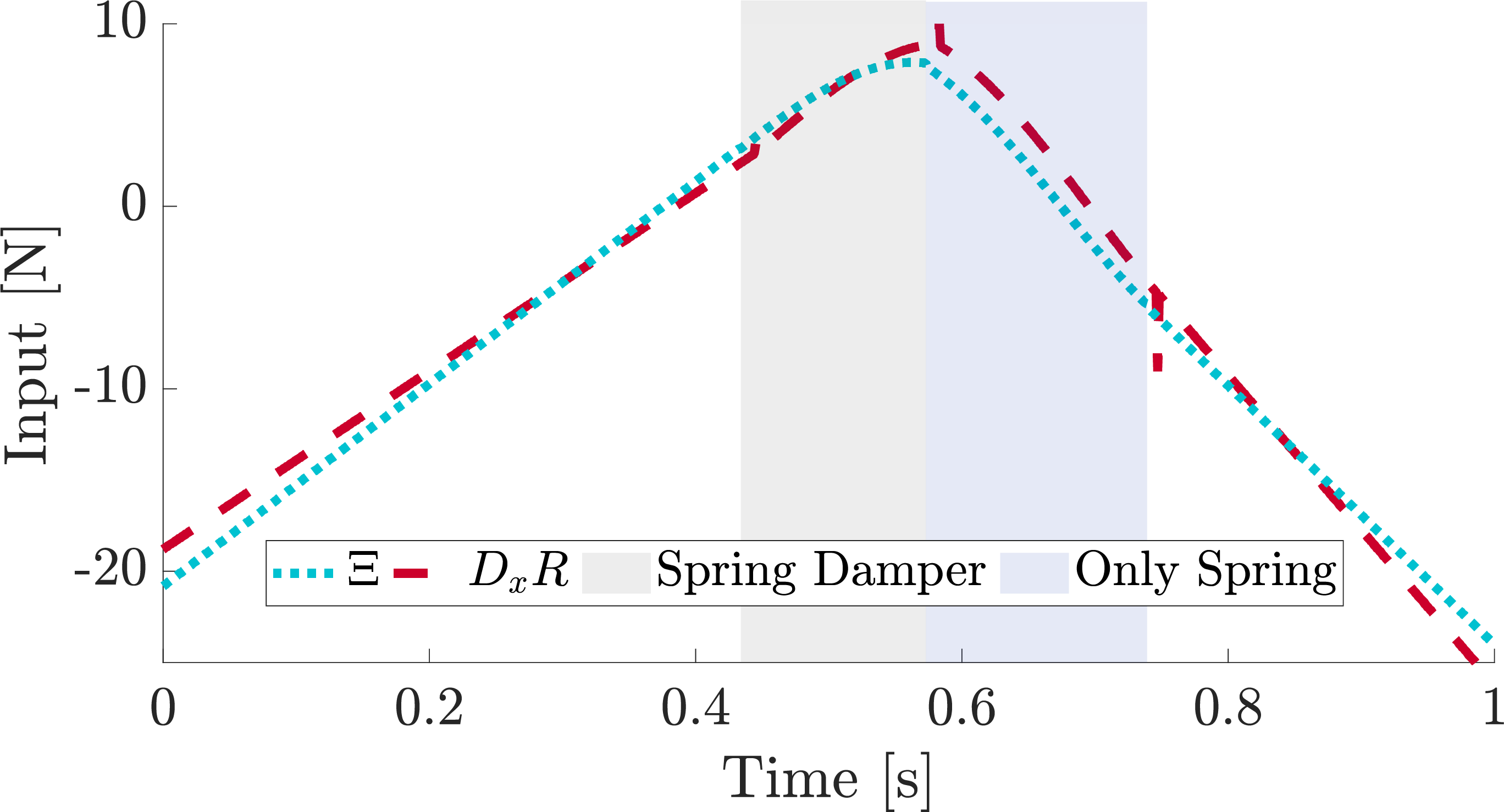}
    \caption{Bouncing ball on a spring-damper ground where both gradient update methods found similar trajectories but using the Jacobian of the reset map $D_xR$ lead to not being able to fully converge as evident by the residual spikes near hybrid transitions. 
    }
    \label{fig:spring_drop_gains}
\end{figure}
For this experiment, $\Xi$-iLQR and $D_xR$-iLQR came up with similar solutions where the cost of $\Xi$-iLQR $J=13.21$ is slightly lower than $D_xR$-iLQR $J=13.29$. This difference is highlighted in Fig. \ref{fig:spring_drop_gains} where $D_xR$-iLQR was not able to smooth out the spikes near mode changes. 
This is also reflected in $D_xR$-iLQR having a higher expected cost reduction as well $\delta J = 0.001$ where $\Xi$-iLQR is a magnitude lower $\delta J = 0.00017$. 
This difference in convergence can most likely be attributed to $D_xR$ providing gradient information that does not adjust the input pre-impact accordingly to allow for adjustments on the spikes post-impact without destructively changing the resulting end state.
\subsection{Ball drop on a curved surface with plastic impacts}
    \begin{figure}[t]
    \centering
    \vspace{0.5em}
    \includegraphics[width = \columnwidth]{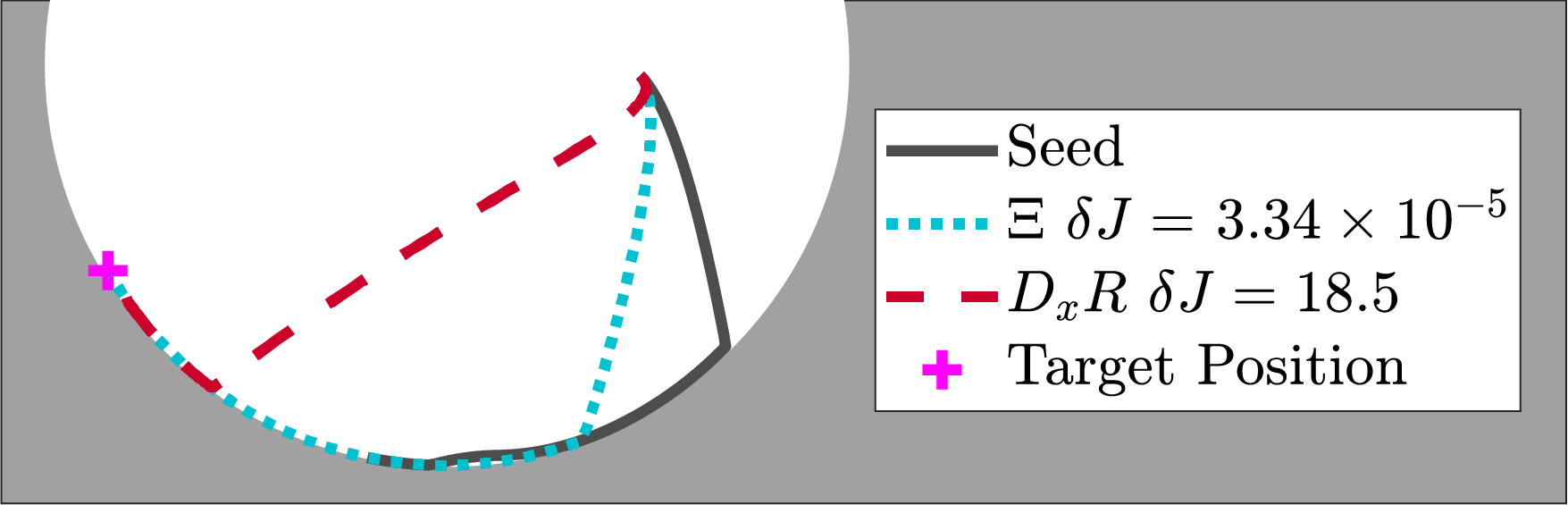}
    \caption{Ball drop on a curved surface with plastic impacts where both gradient methods produced trajectories that got to the end goal, but using $D_xR$ did not converge and had a significantly higher cost. 
    }
    \label{fig:circle_drop_fig}
\end{figure}
The trajectory produced by $\Xi$-iLQR has a cost of $J=10.7$ and $D_xR$-iLQR a cost of $J=50.5$.
The generated position trajectories along with the initial seeded trajectory are shown in Fig. \ref{fig:circle_drop_fig} where both methods ended up at the goal state but $D_xR$-iLQR converged significantly less than $\Xi$-iLQR.

In this example, we purposely seeded a sub-optimal trajectory which releases the contact for a small duration and returns back to the constraint to evaluate if the algorithms would modify the contact sequence. 
$\Xi$-iLQR ended up removing this erroneous contact change and whereas $D_xR$-iLQR ended up not going back to the constraint surface and ended in the unconstrained mode.
We speculate that because $D_xR$ has the wrong gradient information about contacts, it ended up staying in the unconstrained mode for a longer duration and ultimately could not converge.


\subsection{Perching quadcopter}
In this example, the final position trajectories are shown in Fig. \ref{fig:quadcopter_fig} where $\Xi$-iLQR converged $\delta J = 0.170$ with a cost of $J = 4.76$ whereas $D_xR$-iLQR did not converge $\delta J = 3\times10^5$ and produced an erratic solution with very high cost of $J = 2.66\times10^3$. 

$\Xi$-iLQR seemed to make the natural extension of the seed and followed the constraint until the target position was achieved, but removed the double constrained mode at the end. 
We postulate that the fully constrained mode was removed in order to better fine tune the final position because position error is weighted significantly more than velocity.
However, the true optimal solution should include the fully constrained mode to eliminate any velocity for free.
\section{Discussion}
In this work, we extended iLQR to hybrid dynamical systems with piecewise smooth solutions with state jumps.
We compared our algorithm ($\Xi$-iLQR) against using the incorrect hybrid backwards pass update ($D_xR$-iLQR) over a variety of hybrid systems.
For each example, $\Xi$-iLQR outperformed $D_xR$-iLQR in terms of cost and convergence when there was a hybrid transition in the final trajectory.
This result is expected because the saltation matrix is the correct linearization about a hybrid transition.

We believe that our algorithm excels at refining a trajectory which has an initial hybrid mode sequence that needs the timing to be refined.
This is similar to other shooting methods, where they are sensitive to initialization.
However, this issue of locality is accentuated in our algorithm by only giving gradient information and control reference for transitions it has seen.

In future work, we will investigate  systems with intersecting hybrid guards where the Bouligand derivative \cite {burden2016event,scholtes2012introduction} will play an analogous role as the saltation matrix.

\bibliographystyle{IEEEtran}
\bibliography{references}
\end{document}